\documentclass[final]{siamltex}

\usepackage[parfill]{parskip}    
\usepackage{amsmath,amssymb,latexsym,enumerate,mathrsfs}
\usepackage{hyperref}
\usepackage{array}
\usepackage{times}
\usepackage[]{mdframed}
\usepackage{epstopdf}
\usepackage{subfig}
\usepackage{graphicx}
\usepackage{hyperref} 
\usepackage{multicol}
\usepackage{framed}
\usepackage{moreverb}
\usepackage{marvosym}
\usepackage{color,soul}
\definecolor{lightblue}{rgb}{.90,.95,1}
\sethlcolor{yellow}

\title{Simulation-based Bayesian inference with ameliorative learned summary statistics -- Part I}

\author{Getachew K. Befekadu\footnote{\scriptsize Getachew K. Befekadu is with the Department of Electrical \& Computer Engineering, College of Engineering, Physics, and Computing, The Catholic University of America, ~ Washington, DC 20064, USA. E-mail:\,{\tt befekadu@cua.edu}}}

\begin{document}
\maketitle

\renewcommand{\thefootnote}{\arabic{footnote}}

\begin{abstract}
This paper, which is Part~1 of a two-part paper series, considers a simulation-based inference with learned summary statistics, in which such a learned summary statistic serves as an empirical-likelihood with ameliorative effects in the Bayesian setting, when the exact likelihood function associated with the observation data and the simulation model is difficult to obtain in a closed form or computationally intractable. In particular, a transformation technique which leverages the Cressie-Read discrepancy criterion under moment restrictions is used for summarizing the learned statistics between the observation data and the simulation outputs, while preserving the statistical power of the inference. Here, such a transformation of data-to-learned summary statistics also allows the simulation outputs to be conditioned on the observation data, so that the inference task can be performed over certain sample sets of the observation data that are considered as an empirical relevance or believed to be particular importance. Moreover, the simulation-based inference framework discussed in this paper can be extended further, and thus handling weakly dependent observation data. Finally, we remark that such an inference framework is suitable for implementation in distributed computing, i.e., computational tasks involving both the data-to-learned summary statistics and the Bayesian inferencing problem can be posed as a unified distributed inference problem that will exploit distributed optimization and MCMC algorithms for supporting large datasets associated with complex simulation models.
\end{abstract}

\section{Introduction} \label{S1}
In this paper, which is Part~1 of a two-part paper series, we considers a simulation-based inference with learned summary statistics, in which such a learned summary statistic serves as an empirical-likelihood with ameliorative effects in the Bayesian setting, when the exact likelihood function associated with the observation data and the simulation model is difficult to obtain in a closed form or computationally intractable. In particular, we use a transformation technique that leverages the Cressie-Read discrepancy measures under moment (and also conditional) restrictions for summarizing the learned statistics between the observation data and the simulation outputs, while preserving the statistical power of the inference. Moreover, such a transformation of data-to-learned summary statistics allows the simulation outputs to be conditioned on the observation data, so that the inference task can be performed over certain sample sets of the observation data that are considered as an empirical relevance or believed to be particular importance. In this paper, our intent is to provide a formal problem statement along with step-by-step coherent mathematical arguments that are necessary for the exposition of simulation-based inference with learned summary statistics, rather than considering any specific numerical problem. Some numerical works using the proposed computational frameworks have been done and detailed results will be presented elsewhere (i.e., in Part~2 of this two-part paper series). Furthermore, the simulation-based inference framework discussed in this paper can be extended further, and thus handling weakly dependent observation data. Finally, it is worth remarking that such an inference framework is suitable for implementation in distributed computing, i.e., computational tasks involving both the data-to-learned summary statistics and the Bayesian inferencing problem can be posed as a unified distributed inference problem that will exploit distributed optimization and MCMC algorithms for supporting large datasets associated with complex simulation models.

\subsection*{Motivation} \label{S1.1}
Over the last few years, major advances have occurred in the field of simulation-based inference and, more importantly, researchers as well as practitioners across various domains of engineering and applied science are now heavily relying on complex mathematical and simulation models for interpreting observational output datasets, forecasting system behaviors, and making decisions or optimal experiment design. While these complex simulation models provide useful information with high-fidelity, however due to intractability of the exact likelihoods, these models are also poorly suited for statistical inference problems including assessing the predictive capabilities and robustness or estimating model parameters for inverse problems that are required for rigorous uncertainty quantification on quantities of interest. Hence, developing a viable simulation-based inference, that gives additional momentum to the field of statistical inference and, at the same time, that provides reliable solutions with practical pipeline interface for the uncertainty quantification researchers and practitioners, is necessarily required. The main objective of this paper, which is Part~1 of a two-part paper series, is to provide a simulation-based inference framework -- leveraging algorithmic implementations for summarizing learned statistics in the Bayesian setting -- that preserves the statistical power of the inference for estimating quantities of interest and their uncertainties, interpreting observed simulated output datasets or forecasting system behaviors (e.g., see \cite{r1}-\cite{r7} for related discussions on simulation-based inference). 

The remainder of this paper is organized as follows. In Section~\ref{S2}, we present our main results including formal problem statements, where we provide coherent mathematical arguments that are necessary for the exposition of simulation-based inference with learned summary statistics. The section also presents a transformation technique for data-to-learned summary statistics, based on the Cressie-Read discrepancy measure under moment (and also conditional) moment restrictions, that serve as a proxy to the exact log-likelihood function (i.e., an approximation to the exact intractable log-likelihood) in the Bayesian setting. Moreover, the mathematical formulations presented in this section also provide a viable way for algorithmic implementations in distributed computing that will support large datasets associated with complex simulation models. Finally, Section~\ref{S3} contains some concluding remarks

\section{Main results} \label{S2}
\begin{figure}[hbt]
\begin{center}
 \includegraphics[scale=0.5]{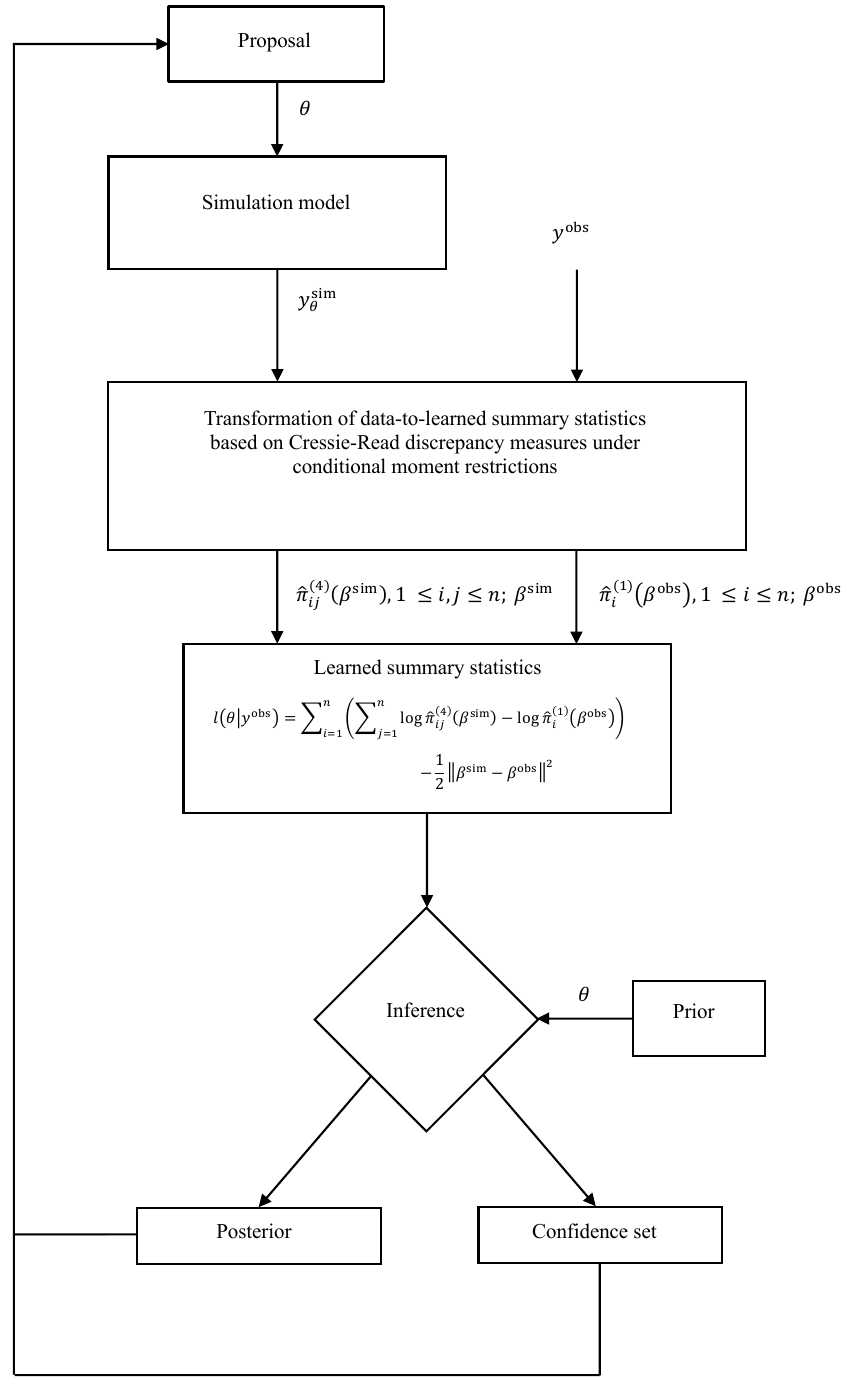}
 \caption{\small Inference under conditional moment restrictions.} \label{Figure:FG3}
\end{center}
\end{figure}

In this section, we present our main results that include formal problem statements, in which we provide step-by-step coherent mathematical arguments that are necessary for the exposition of simulation-based inference with learned summary statistics (see Fig.~\ref{Figure:FG1},\, Fig.~\ref{Figure:FG2} \,and Fig.~\ref{Figure:FG3}). Here, we stress that such data-to-learned summary statistics, which preserve the statistical power of the inference, will serve as a proxy to the exact log-likelihood function (i.e., an approximation to the exact intractable log-likelihood) in the Bayesian setting. Moreover, the mathematical formulations presented in this section provide a viable way for algorithmic implementations in distributed computing that will support large datasets associated with complex simulation models.
\subsection{Inference with learned summary statistics - Cressie-Read discrepancy measures under moment restrictions} \label{S2.1}
\begin{figure}[h]
\begin{center}
 \includegraphics[scale=0.5]{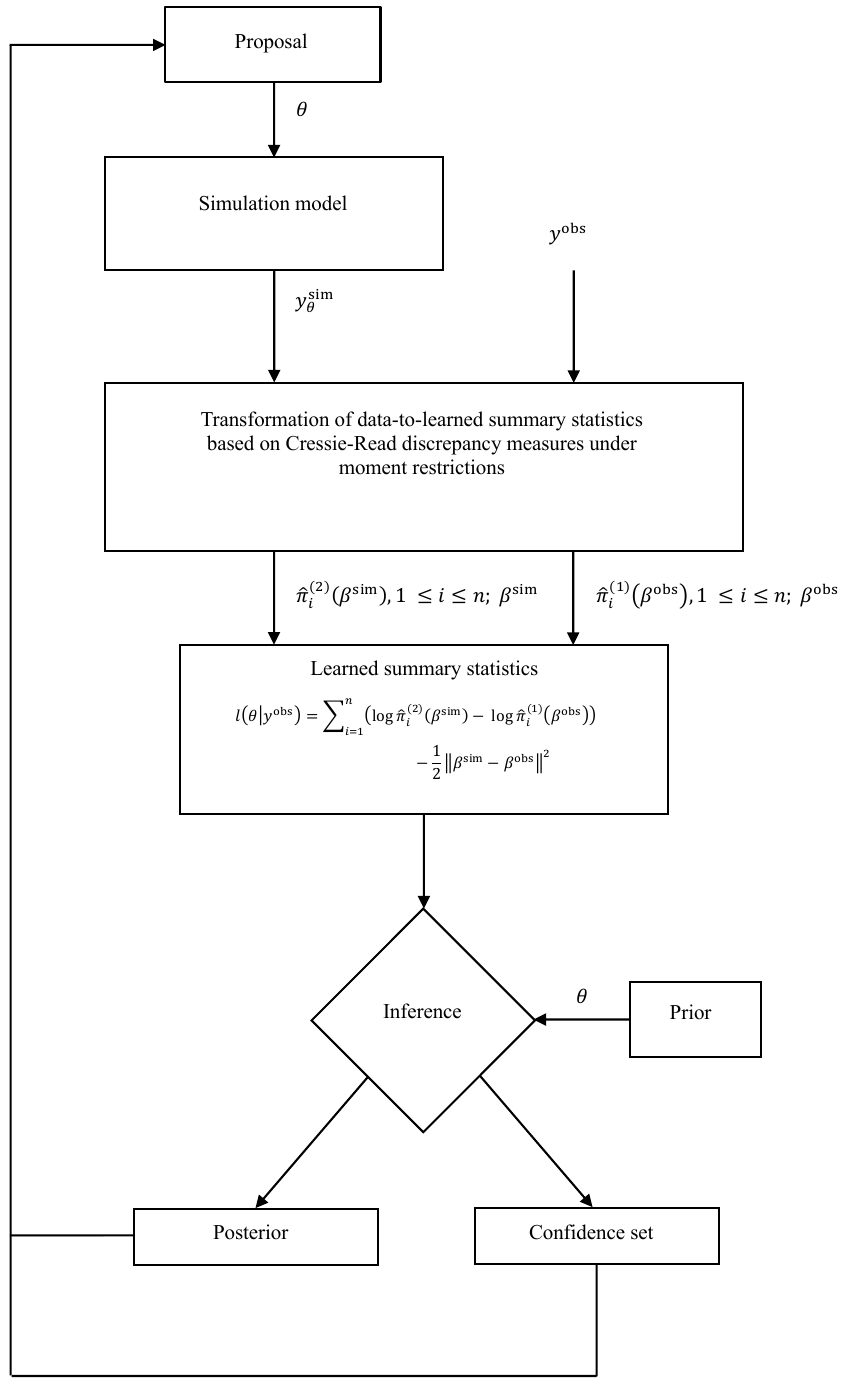}
 \caption{\small Inference with data-to-learned summary statistics.} \label{Figure:FG1}
\end{center}
\end{figure}

Suppose we have independent observation data $y^{\rm obs} =\bigl\{y_i^{\rm obs} \bigr\}_{i=1}^n$, with support $\mathbb{R}^{d_y}$ (see also Fig.~\ref{Figure:FG1}). Then, we consider the following Cressie-Read discrepancy criterion of the form
\begin{align}
\min_{\pi_i, 1\le i \le n;\, \beta^{\rm obs}} \, \frac{1}{\gamma(\gamma +1)} \sum\nolimits_{i=1}^n \left[\bigl(n \pi_i \bigr)^{\gamma +1} - 1 \right], \quad \text{with} \quad \gamma \in \mathbb{R}\setminus\{-1,0\}, \label{Eq2.1}
\end{align}
where the above minimization is carried out under the following restrictions
\begin{enumerate}[(i)]
\item $\sum\nolimits_{i=1}^n \pi_i g\bigl(y_i^{\rm obs}, \beta^{\rm obs} \bigr) = 0$,
\item $\sum\nolimits_{i=1}^n \pi_i = 1$, with $\pi_i \ge 1$, for $i=1$, $2$, \ldots, $n$,
\end{enumerate}
while $g \colon \mathbb{R}^{d_y} \times \mathbb{B} \to \mathbb{R}^{d_g}$, with $\mathbb{B} \subseteq \mathbb{R}^{d_{\beta}}$, is a measurable function for modeling moment restrictions, i.e., $E\left[ g\bigl(y, \beta^{\rm obs}\bigr)\right] = 0$, and $\beta^{\rm obs} \in \mathbb{B}$ is a vector of unknown parameters that satisfies the above moment restrictions.\footnote{The expression $\frac{1}{\gamma(\gamma +1)} \sum\nolimits_{i=1}^n \bigl[\bigl(n \pi_i \bigr)^{\gamma +1} - 1 \bigr]$ is usually interpreted as limiting cases for $\gamma = -1$ or $0$.} Moreover, for a given $\gamma$, the above minimum discrepancy estimator for the parameters $\beta^{\rm obs} \in \mathbb{B}$ minimizes with respect to $\pi_i$, for $i=1$, $2$, \ldots, $n$, while the parameters $\beta^{\rm obs}$ are restricted to satisfy the above two constraints, i.e., $\sum\nolimits_{i=1}^n \pi_i g\bigl(y_i^{\rm obs}, \beta^{\rm obs} \bigr) = 0$ and $\sum\nolimits_{i=1}^n \pi_i = 1$, with $\pi_i \ge 1$, for $i=1$, $2$, \ldots, $n$. Here, we remark that the above Cressie-Read discrepancy criterion makes an exact comparison between the probabilities $\bigl\{\pi_i\bigr\}_{i=1}^n$, that take account of the moment restrictions $\sum\nolimits_{i=1}^n \pi_i g\bigl(y_i^{\rm obs}, \beta^{\rm obs} \bigr) = 0$, and that of the unconstrained empirical distribution function counterparts $\pi_i=1/n$, for $i=1$, $2$, \ldots, $n$, that solve the minimization problem in the absence of moment restrictions (e.g., see \cite{r8}, \cite{r9}, \cite{r10} and \cite{r11} for some discussions based on the work of \cite{r12}; see also \cite{r13}, \cite{r14} and \cite{r15} for related discussions on empirical likelihood methods with moment restrictions).\footnote{Note that the Cressie-Read discrepancy in Equation~\eqref{Eq2.1} is a particular form of entropy known as the Renyi's $\alpha$-class of generalized measures of entropy (see \cite{r16}).} Later, such a mathematical argument will be exploited as a transformation technique for summarizing data-to-leaned statistics between the observation data and that of the simulation outputs that will also allow us to explore the learned summary statistics using the MCMC algorithm in the Bayesian setting (see Figures~\ref{Figure:FG1}).

With a standard Lagrangian argument, the optimization problem in Equation~\ref{Eq2.1} can then be conveniently solved by using Lagrange multipliers (i.e., based on the KKT's optimality conditions  (e.g., see \cite{r17})) as follows
\begin{align}
\mathscr{L}\bigl(\beta^{\rm obs}, \pi, \lambda, \eta  \bigr) =& \frac{1}{\gamma(\gamma +1)} \sum\nolimits_{i=1}^n \left[\bigl(n \pi_i \bigr)^{\gamma +1} - 1 \right] \notag\\
& \quad -  \lambda^T\sum\nolimits_{i=1}^n \pi_i g\bigl(y_i^{\rm obs}, \beta^{\rm obs} \bigr) - \eta \left(\sum\nolimits_{i=1}^n \pi_i - 1 \right), \label{Eq2.2}
\end{align}
where $\lambda \in \mathbb{R}^{d_g}$ and $\eta \in \mathbb{R}$ are Lagrange multipliers associated with the restrictions in the optimization problem of Equation~\eqref{Eq2.1}. For a fixed $\beta^{\rm obs} \in \mathbb{B}$, using the envelop theorem, it is easy to see that the solutions for $\hat{\pi}_i^{(1)}(\beta^{\rm obs})$, for $i = 1$, $2$, \ldots, $n$, satisfy the following system of equations
\begin{align}
\frac{n}{\gamma} \bigl(n \hat{\pi}_i^{(1)}(\beta^{\rm obs}) \bigr)^{\gamma} - \hat{\lambda}(\beta^{\rm obs})^T g\bigl(y_i^{\rm obs}, \beta^{\rm obs} \bigr) - \hat{\eta}(\beta^{\rm obs}) = 0, \label{Eq2.3}
\end{align}
where $\hat{\lambda}(\beta^{\rm obs})$ and $\hat{\eta}(\beta^{\rm obs})$ are, respectively, the Lagrange multipliers associated with the constraints 
\begin{align*}
\sum\nolimits_{i=1}^n \hat{\pi}_i^{(1)}(\beta^{\rm obs}) g\bigl(y_i^{\rm obs}, \beta^{\rm obs} \bigr) = 0 \quad \text{and} \quad \sum\nolimits_{i=1}^n \hat{\pi}_i^{(1)}(\beta^{\rm obs}) = 1.
\end{align*}
Moreover, the solutions $\hat{\pi}_i^{(1)}(\beta^{\rm obs})$, for $i = 1$, $2$, \ldots, $n$, are given in a closed form as follows
\begin{align}
\hat{\pi}_i^{(1)}(\beta^{\rm obs}) = \left[ \frac{\gamma}{n^{\gamma +1}} \left( \hat{\lambda}(\beta^{\rm obs})^T g\bigl(y_i^{\rm obs}, \beta^{\rm obs} \bigr) + \hat{\eta}(\beta^{\rm obs}) \right)\right]^{\frac{1}{\gamma}}. \label{Eq2.4}
\end{align}
Note that
\begin{align*}
\sum\nolimits_{i=1}^n \hat{\pi}_i^{(1)}\bigl(\beta^{\rm obs}\bigr) &= \sum\nolimits_{i=1}^n\left[ \frac{\gamma}{n^{\gamma +1}} \left( \hat{\lambda}\bigl(\beta^{\rm obs}\bigr)^T g\bigl(y_i^{\rm obs}, \beta^{\rm obs} \bigr) + \hat{\eta}\bigl(\beta^{\rm obs}\bigr) \right)\right]^{\frac{1}{\gamma}}\\
         & = 1.
\end{align*}
 Then, the solutions in Equation~\ref{Eq2.4} are further given by
\begin{align}
\hat{\pi}_i^{(1)}(\beta^{\rm obs}) = \frac{\left(\hat{\lambda}(\beta^{\rm obs})^T g\bigl(y_i^{\rm obs}, \beta^{\rm obs} \bigr) + \hat{\eta}(\beta^{\rm obs}) \right)^{\frac{1}{\gamma}}}{ \sum\nolimits_{k=1}^n \left(\hat{\lambda}(\beta^{\rm obs})^T g\bigl(y_k^{\rm obs}, \beta^{\rm obs} \bigr) + \hat{\eta}(\beta^{\rm obs}) \right)^{\frac{1}{\gamma}}}, \quad i=1, 2, \ldots, n. \label{Eq2.5}
\end{align}
With minor abuse of notation, i.e., if we take $\hat{\eta}(\beta^{\rm obs})$ as a common factor from both the numerator and denominator of the above equation and then replacing ${\hat{\lambda}(\beta^{\rm obs})}\bigl/{\hat{\eta}(\beta^{\rm obs})}$  with $\hat{\lambda}(\beta^{\rm obs})$, then we will have the following equation 
\begin{align}
\hat{\pi}_i^{(1)}(\beta^{\rm obs}) = \frac{\left(1+ \hat{\lambda}(\beta^{\rm obs})^T g\bigl(y_i^{\rm obs}, \beta^{\rm obs} \bigr) \right)^{\frac{1}{\gamma}}}{ \sum\nolimits_{k=1}^n \left(1+ \hat{\lambda}(\beta^{\rm obs})^T g\bigl(y_k^{\rm obs}, \beta^{\rm obs} \bigr)\right)^{\frac{1}{\gamma}}}, \quad i=1, 2, \ldots, n. \label{Eq2.6}
\end{align}
On the other hand, if we consider the simulation outputs $y_{\theta}^{\rm sim} =\bigl\{y_{\theta,i}^{\rm sim} \bigr\}_{i=1}^n$, with support $\mathbb{R}^{d_y}$, where we emphasized such simulation outputs are also dependent on the simulation model $\theta$, with support $\Theta \subset \mathbb{R}^{d_{\theta}}$. Then, the corresponding optimization problem - which is based on the Cressie-Read discrepancy criterion - has the following form
\begin{align}
\min_{\pi_i, 1\le i \le n;\, \beta^{\rm sim}} \, \frac{1}{\gamma(\gamma +1)} \sum\nolimits_{i=1}^n \left[\bigl(n \pi_i \bigr)^{\gamma +1} - 1 \right], \label{Eq2.7}
\end{align}
where the above minimization is also carried out under the following restrictions
\begin{enumerate}[(i)]
\item $\sum\nolimits_{i=1}^n \pi_i g\bigl(y_{\theta,i}^{\rm sim}, \beta^{\rm sim} \bigr) = 0$,
\item $\sum\nolimits_{i=1}^n \pi_i = 1$, with $\pi_i \ge 1$, for $i=1$, $2$, \ldots, $n$.
\end{enumerate}
Note that, for a fixed $\beta^{\rm sim} \in \mathbb{B}$, the corresponding solutions for $\hat{\pi}_i^{(2)}(\beta^{\rm sim})$ satisfy the following system of equations
\begin{align}
\frac{n}{\gamma} \bigl(n \hat{\pi}_i^{(2)}(\beta^{\rm sim}) \bigr)^{\gamma} - \hat{\lambda}(\beta^{\rm sim})^T g\bigl(y_{\theta,i}^{\rm sim}, \beta^{\rm sim} \bigr) - \hat{\eta}(\beta^{\rm sim}) = 0, \quad i = 1, 2, \ldots, n, \label{Eq2.8}
\end{align}
where $\hat{\lambda}(\beta^{\rm sim})$ and $\hat{\eta}(\beta^{\rm sim})$ are, respectively, the Lagrange multipliers associated with the constraints 
\begin{align*}
\sum\nolimits_{i=1}^n \hat{\pi}_i^{(2)}(\beta^{\rm sim}) g\bigl(y_{\theta,i}^{\rm sim}, \beta^{\rm sim} \bigr) = 0 \quad \text{and} \quad \sum\nolimits_{i=1}^n \hat{\pi}_i^{(2)}(\beta^{\rm sim}) = 1.
\end{align*}
Moreover, if we follow the same arguments as above, i.e., replacing ${\hat{\lambda}(\beta^{\rm sim})}\bigl/{\hat{\eta}(\beta^{\rm sim})}$  by $\hat{\lambda}\bigl(\beta^{\rm sim}\bigr)$, then the solutions $\hat{\pi}_i^{(2)}(\beta^{\rm sim})$, for  $i = 1$, $2$, \ldots, $n$, are given by 
\begin{align}
\hat{\pi}_i^{(2)}(\beta^{\rm sim}) = \frac{\left(1+ \hat{\lambda}(\beta^{\rm sim})^T g\bigl(y_{\theta,i}^{\rm sim}, \beta^{\rm sim} \bigr) \right)^{\frac{1}{\gamma}}}{ \sum\nolimits_{k=1}^n \left(1+ \hat{\lambda}(\beta^{\rm sim})^T g\bigl(y_{\theta,k}^{\rm sim}, \beta^{\rm sim} \bigr)\right)^{\frac{1}{\gamma}}}. \label{Eq2.9}
\end{align}
In what follows, we introduce the data-to-learned summary statistics $\ell_{\rm learned}\bigl(\theta \vert y^{\rm obs}\bigr)$ based on:
\begin{enumerate} [(i).]
 \item the empirical log-likelihood ratio between the Cressie-Read discrepancy contrast probabilities $\bigr\{\hat{\pi}_i^{(2)}(\beta^{\rm sim})\bigr\}_{i=1}^n$ and $\bigl\{\hat{\pi}_i^{(1)}(\beta^{\rm obs})\bigr\}_{i=1}^n$, and 
 \item the Euclidean norm distance between the estimated parameters $\beta^{\rm sim}$ and $\beta^{\rm obs}$ that are associated with the moment restrictions,
 \end{enumerate}
that is,
\begin{align}
\ell_{\rm learned}\bigl(\theta \vert y^{\rm obs}\bigr) = \sum\nolimits_{i=1}^n \biggl( \log \hat{\pi}_i^{(2)}(\beta^{\rm sim})  - \log \hat{\pi}_i^{(1)}(\beta^{\rm obs})\biggr) - \frac{1}{2} \bigl\Vert \beta^{\rm sim} - \beta^{\rm obs}\bigr\Vert^2, \label{Eq2.10}
\end{align}
where such a data-to-learned summary statistic (which preserves the statistical power of the inference) can serve as a proxy to the exact log-likelihood function (which is associated with the observation data $y^{\rm obs}$ and the simulation model $\theta$) for inferencing in the Bayesian setting or the conventional maximum-likelihood based estimator. Here, we remark that both optimization problems in Equation~\eqref{Eq2.1} and Equation~\eqref{Eq2.7} can be solved using search algorithms such as the Nelder-Mead optimization algorithm (e.g., see\cite{r18}) that can be implemented in distributed optimization involving parallelizing functions evaluations and updates across multiple computational resources (which is the focus of Part 2 in this two-part paper series).

\subsubsection*{Exploring $\ell_{\rm learned}\bigl(\theta \vert y^{\rm obs}\bigr)$ by the MCMC algorithm}
Suppose that the parameter $\theta$ has a prior distribution $p_{\rm pr}(\theta)$. Then, for a given learned summary statistic $\ell_{\rm learned}\bigl(\theta \vert y^{\rm obs}\bigr)$, which serves as a proxy to the exact log-likelihood function, the posterior density $p(\theta)$ can be determined as follows
\begin{align*}
p(\theta) = \ell_{\rm learned}\bigl(\theta \vert y^{\rm obs}\bigr) p_{\rm pr}(\theta),
\end{align*}
where such a posterior distribution information is useful for estimating some of the quantities of interest and their uncertainty. Here, our main focus is to guide the simulator with parameter points $\theta$ that will increase our knowledge most, i.e., in the Bayesian setting requiring to explore $\ell_{\rm learned}\bigl(\theta \vert y^{\rm obs}\bigr)$ using the MCMC algorithm (e.g., see \cite{r19}, \cite{r20}, \cite{r21} and \cite{r22} for additional discussions):
\begin{enumerate}[(i)]
\item[{\bf ~ }] Starting from a parameter guess $\theta^{\rm (0)} \in \Theta$, repeat the following for $k=1$, $2$, \ldots
\item[{\bf i.}] Propose for $\theta$ such that 
\begin{align*}
\theta^{\rm prop} = \theta^{\rm (k-1)} + \delta^{\rm (k)},
\end{align*}
where $\delta^{\rm (k)}$, with support $\Theta$, is a random vector sampled from a suitable symmetric distribution.
\item[{\bf ii.}] Then, set
\begin{align*}
& \quad \quad \theta^{\rm (k)} = \theta^{\rm prop} \quad \text{with probability}\\
 & \quad\quad\quad  \min \left\{1, \, \exp \bigl(\ell_{\rm learned}\bigl(\theta^{\rm prop} \vert y^{\rm obs}\bigr) - \ell_{\rm learned}\bigl(\theta^{\rm (k-1)} \vert y^{\rm obs}\bigr) \bigr) \right\},\\
 &\quad \quad \theta^{\rm (k)} = \theta^{\rm (k-1)} \quad \text{otherwise}.
\end{align*}
\item[{\bf iii.}] Until convergence, i.e.,
$\bigl\Vert \ell_{\rm learned}\bigl(\theta^{\rm (k)} \vert y^{\rm obs}\bigr) - \ell_{\rm learned}\bigl(\theta^{\rm (k-1)} \vert y^{\rm obs}\bigr)  \bigr\Vert \le \epsilon_{\rm tol}$, for some error tolerance $\epsilon_{\rm tol} > 0$.
\end{enumerate} 
Here, it is worth remarking that a hierarchical formulation of exploring the learned statistics $\ell_{\rm learned}\bigl(\theta \vert y^{\rm obs}\bigr)$ is often advantageous as to provide fairly a set of hierarchical, but simpler, inference problems that otherwise would be computationally impossible.

\subsection{Inference with multiple replication of simulation outputs} \label{S2.2}
\begin{figure}[h]
\begin{center}
 \includegraphics[scale=0.5]{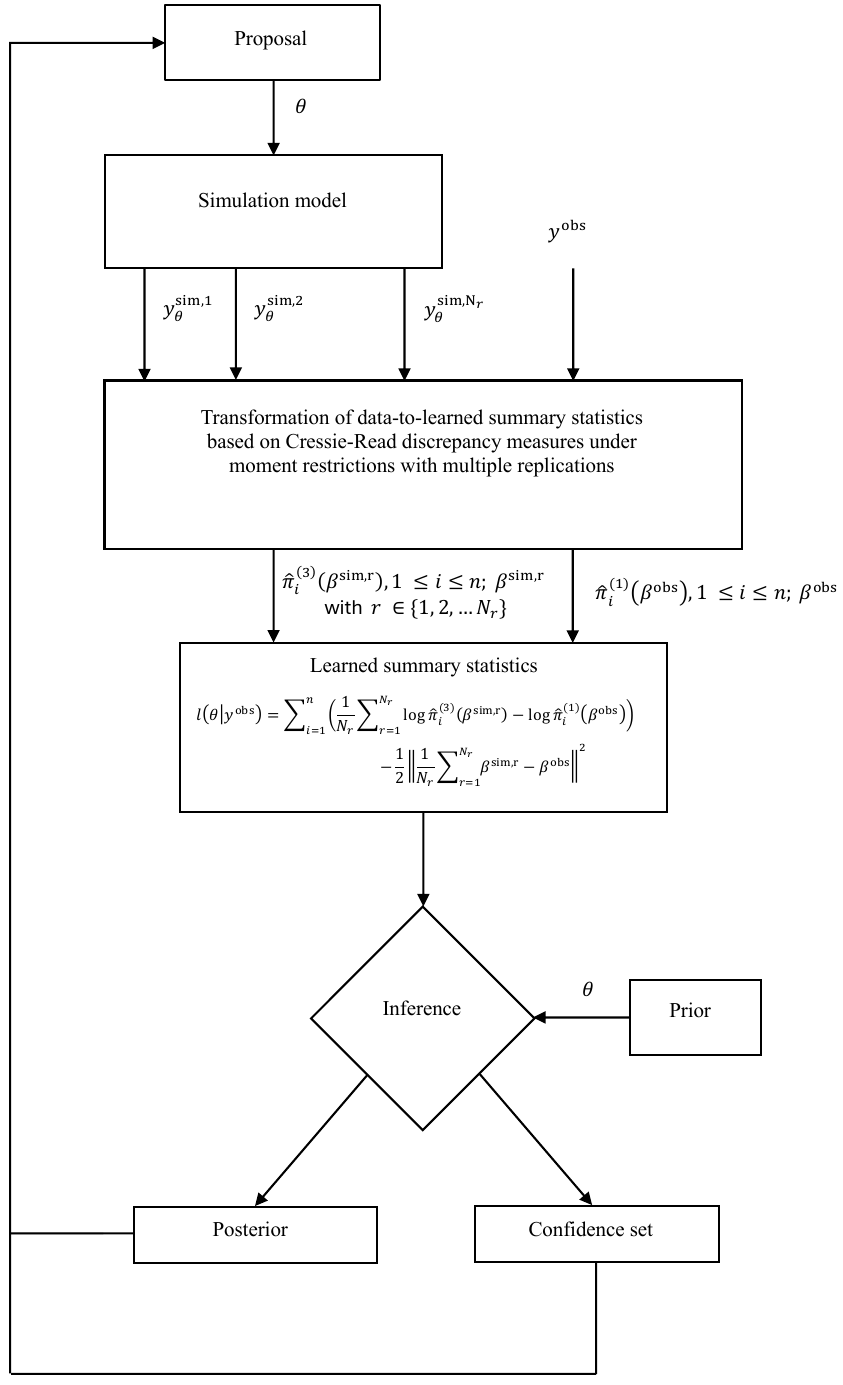}
 \caption{\small Inference with multiple replication of simulation outputs.} \label{Figure:FG2}
\end{center}
\end{figure}

Suppose that we have a multiple-run of independent simulation output replications $y_{\theta}^{\rm sim, 1}$, $y_{\theta}^{\rm sim, 2}$, \ldots, $y_{\theta}^{\rm sim, N_r}$, with  $y_{\theta}^{\rm sim, r}=\bigl\{y_{\theta,i}^{\rm sim, r} \bigr\}_{i=1}^n$, for $r \in \{1,2, \ldots, N_r\}$ (see also Fig.~\ref{Figure:FG2}). Then, we consider a family of optimization problems - based on the Cressie-Read discrepancy criterion - of the form
\begin{align}
\min_{\pi_i, 1\le i \le n; \, \beta^{\rm sim,r}} &\, \frac{1}{\gamma(\gamma +1)} \sum\nolimits_{i=1}^n \left[\bigl(n \pi_i \bigr)^{\gamma +1} - 1 \right], \label{Eq2.11}\\
&\text{s.t.} \notag\\
& \quad \sum\nolimits_{i=1}^n \pi_i g\bigl(y_{\theta,i}^{\rm sim,r}, \beta^{\rm sim,r} \bigr) = 0, \notag\\
& \quad \sum\nolimits_{i=1}^n \pi_i = 1, \quad \text{with}\quad \pi_i \ge 1, \quad i=1, 2, \ldots, n, \notag
\end{align}
for $r \in \{1,\,2, \dots, N_r\}$. 

Note that, if we follow the same steps as in Subsection~\ref{S2.1} and working with the Lagrange multipliers associated with the constraints in Equation~\ref{Eq2.11}, i.e., replacing ${\hat{\lambda}\bigl(\beta^{\rm sim,r}\bigr)}\bigl/{\hat{\eta}\bigl(\beta^{\rm sim,r}\bigr)}$ by $\hat{\lambda}\bigl(\beta^{\rm sim,r}\bigr)$ for each $r \in \{1,\,2, \dots, N_r\}$. Then, it is easy to see that the solutions $\hat{\pi}_i^{(3)}(\beta^{\rm sim,r})$, for each $r \in \{1,\,2, \dots, N_r\}$, are given by
\begin{align}
\hat{\pi}_i^{(3)}(\beta^{\rm sim,r}) = \frac{\left(1+ \hat{\lambda}(\beta^{\rm sim,r})^T g\bigl(y_{\theta,i}^{\rm sim,r}, \beta^{\rm sim,r} \bigr) \right)^{\frac{1}{\gamma}}}{ \sum\nolimits_{k=1}^n \left(1+ \hat{\lambda}(\beta^{\rm sim,r})^T g\bigl(y_{\theta,k}^{\rm sim,r}, \beta^{\rm sim,r} \bigr)\right)^{\frac{1}{\gamma}}}, \quad i = 1, 2, \ldots, n. \label{Eq2.12}
\end{align}

Moreover, we introduce the data-to-learned summary statistics $\ell_{\rm learned}\bigl(\theta \vert y^{\rm obs}\bigr)$ based on:
\begin{enumerate} [(i).]
 \item the empirical log-likelihood ratio between the averaged Cressie-Read discrepancy contrast probabilities $\bigl\{\frac{1}{N_r} \sum\nolimits_{i=1}^{N_r} \hat{\pi}_i^{(3)}(\beta^{\rm sim,r})\bigr\}_{i=1}^n$ and $\bigl\{\hat{\pi}_i^{(1)}(\beta^{\rm obs})\bigr\}_{i=1}^n$, and
 \item the Euclidean norm distance between the averaged estimated parameters $\frac{1}{N_r} \sum\nolimits_{i=1}^{N_r}\beta^{\rm sim,r}$ and $\beta^{\rm obs}$ that are associated with the moment restrictions,
 \end{enumerate} 
 that is,
\begin{align}
\ell_{\rm learned}\bigl(\theta \vert y^{\rm obs}\bigr) =& \sum\nolimits_{i=1}^n \biggl( \log \biggl(\frac{1}{N_r}\sum\nolimits_{r=1}^{N_r} \hat{\pi}_i^{(3)}(\beta^{\rm sim,r})\biggr)  - \log \hat{\pi}_i^{(1)}(\beta^{\rm obs})\biggr) \notag\\
& \quad\quad\quad - \frac{1}{2} \biggl\Vert \frac{1}{N_r}\sum\nolimits_{r=1}^{N_r} \beta^{\rm sim,r} - \beta^{\rm obs}\biggr\Vert^2, \label{Eq2.13}
\end{align}
where such a learned summary statistic will serve as an approximation to the exact log-likelihood function for the Bayesian inference.

\subsection{Inference under conditional moment restrictions}
In this subsection, we present a transformation technique of data-to-learned summary statistics under conditional moment restrictions that will allow us to perform inference task over certain sample sets of the observation data considered as an empirical relevance or believed to be particular importance. In particular, we consider a local Cressie-Read discrepancy criterion for the optimization problem under conditional moment restrictions of the form
\begin{align}
E\left[ g\bigl(y_{\theta}^{\rm sim}, \beta \bigr) \bigl \vert y^{\rm obs} \right] = 0, \label{Eq2.14}
\end{align}
where $\beta^{\rm sim} \in \mathbb{B}$ is a vector of unknown parameters (see also Equation~\eqref{Eq2.7}). That is, we introduce the following weights $w_{ij}$, for $1\le i,j \le n$, with respect to the observation data $y^{\rm obs} =\bigl\{y_i^{\rm obs} \bigr\}_{i=1}^n$ as follows
\begin{align*}
w_{ij} = {\psi\bigl(\frac{y_i^{\rm obs} - y_j^{\rm obs}}{h}\bigr)}\biggl /{\sum\nolimits_{k=1}^n \psi\bigl(\frac{y_i^{\rm obs}  - y_k^{\rm obs} }{h}\bigr)}, \quad  1 \le i,j \le n,
\end{align*}
for some appropriately chosen kernel function $\psi$ and bandwidth $h$. Then, we have the following optimization problem
\begin{align}
\min_{\pi_{ij}, 1\le i,j \le n; \, \beta^{\rm sim}} &\, \frac{1}{\gamma(\gamma +1)} \sum\nolimits_{i=1}^n \sum\nolimits_{j=1}^n \left[w_{ij}\biggl(\frac{\pi_{ij}} {w_{ij}} \biggr)^{\gamma +1} - 1 \right], \label{Eq2.15} \\
&\text{s.t.} \notag\\
& \quad \sum\nolimits_{j=1}^n \pi_{ij} g\bigl(y_{\theta,j}^{\rm sim}, \beta^{\rm sim} \bigr) = 0, \notag\\
& \quad \sum\nolimits_{j=1}^n \pi_{ij} = 1, \quad i=1, 2, \ldots, n, \notag
\end{align}
with $\pi_{ij} \ge 1$ for $1 \le i,j \le n$. Then, with a standard Lagrangian argument, the above optimization problem can be conveniently solved by using Lagrange multipliers as follows
\begin{align}
\mathscr{L}\bigl(\beta^{\rm sim}, \pi, \lambda, \eta  \bigr) =& \frac{1}{\gamma(\gamma +1)} \sum\nolimits_{i=1}^n \sum\nolimits_{j=1}^n \left[w_{ij}\biggl(\frac{\pi_{ij}} {w_{ij}} \biggr)^{\gamma +1} - 1 \right] \notag\\
& \quad -  \sum\nolimits_{i=1}^n \lambda_i^T\sum\nolimits_{j=1}^n \pi_{ij} g\bigl(y_j^{\rm sim}, \beta^{\rm sim} \bigr) - \sum\nolimits_{i=1}^n \eta \left(\sum\nolimits_{j=1}^n \pi_{ij} - 1 \right), \label{Eq2.16}
\end{align}
where $\lambda_i \in \mathbb{R}^{d_g}$, for $i=1$, $2$, \ldots, $n$, and $\eta \in \mathbb{R}$ (with a common $\eta$ for $i=1$, $2$, \ldots, $n$) are Lagrange multipliers that are associated with the constraints in Equation~\eqref{Eq2.15}. For a fixed $\beta^{\rm sim} \in \mathbb{B}$, using the envelop theorem, then it is easy to see that the solutions for $\hat{\pi}_{ij}^{(4)}(\beta^{\rm sim})$, for $1 \le i,j \le n$, satisfy the following system of equations
\begin{align}
\frac{1}{\gamma} \biggl(\frac{\hat{\pi}_{ij}^{(4)}(\beta^{\rm sim})}{w_{ij}}\biggr)^{\gamma} - \hat{\lambda}_i(\beta^{\rm sim})^T g\bigl(y_j^{\rm sim}, \beta^{\rm sim} \bigr) - \hat{\eta}(\beta^{\rm sim}) = 0, \quad 1 \le i,j \le n, \label{Eq2.17}
\end{align}
where $\hat{\lambda}_i(\beta^{\rm sim})$, for $i=1$, $2$, \ldots, $n$, and $\hat{\eta}(\beta^{\rm sim})$ are, respectively, the Lagrange multipliers associated with the constraints 
\begin{align*}
\sum\nolimits_{j=1}^n \hat{\pi}_{ij}^{(4)}(\beta^{\rm sim}) g\bigl(y_j^{\rm sim}, \beta^{\rm sim} \bigr) = 0 \quad \text{and} \quad \sum\nolimits_{j=1}^n \hat{\pi}_{ij}^{(4)}(\beta^{\rm sim}) = 1.
\end{align*}
Moreover, if we solve for $\hat{\pi}_{ij}^{(4)}(\beta^{\rm sim})$, then we will have following equations
\begin{align}
\hat{\pi}_{ij}^{(4)}(\beta^{\rm sim}) = w_{ij} \left [ \gamma \bigl(\hat{\eta}(\beta^{\rm sim})  + \hat{\lambda}_i(\beta^{\rm sim})^T g\bigl(y_j^{\rm sim}, \beta^{\rm sim} \bigr) \bigr) \right], \quad 1 \le i,j \le n, \label{Eq2.18}
\end{align}
Note that
\begin{align*}
\sum\nolimits_{j=1}^n \hat{\pi}_{ij}^{(4)}(\beta^{\rm sim}) & = \sum\nolimits_{j=1}^n w_{ij} \left [ \gamma \bigl( \hat{\eta}(\beta^{\rm sim})  + \hat{\lambda}_i(\beta^{\rm sim})^T g\bigl(y_j^{\rm sim}, \beta^{\rm sim} \bigr) \bigr) \right]\\
&=1, \quad\quad\quad\quad i=1,2, \ldots, n,
\end{align*}
and if we further replace $\hat{\lambda}_i(\beta^{\rm sim}) \bigl/ \hat{\eta}(\beta^{\rm sim})$ by $\hat{\lambda}_i(\beta^{\rm sim})$, for each $i=1$, $2$, \ldots, $n$. Then, we will have 
following equations
\begin{align}
\hat{\pi}_{ij}^{(4)}(\beta^{\rm sim}) = \frac{w_{ij} \left [ 1 + \hat{\lambda}_i(\beta^{\rm sim})^T g\bigl(y_j^{\rm sim}, \beta^{\rm sim} \bigr) \right]}{\sum\nolimits_{k=1}^n w_{ik} \left [ 1  + \hat{\lambda}_i(\beta^{\rm sim})^T g\bigl(y_k^{\rm sim}, \beta^{\rm sim} \bigr) \right]}, \quad 1 \le i,j \le n. \label{Eq2.19}
\end{align}

Moreover, we can introduce data-to-learned summary statistics $\ell_{\rm learned}\bigl(\theta \vert y^{\rm obs}\bigr)$ based on:
\begin{enumerate} [(i).]
\item the empirical log-likelihood ratio between the local Cressie-Read discrepancy contrast probabilities $\hat{\pi}_{ij}^{(4)}(\beta^{\rm sim})$, for $1 \le i,j \le n$, and $\bigl\{\hat{\pi}_i^{(1)}\bigl(\beta^{\rm obs}\bigr)\bigr\}_{i=1}^n$, and
\item the Euclidean norm distance between the estimated parameters $\beta^{\rm sim}$ and $\beta^{\rm obs}$ that are associated with the conditional moment restrictions,
\end{enumerate}
that is, 
\begin{align}
\ell_{\rm learned}\bigl(\theta \vert y^{\rm obs}\bigr) =& \sum\nolimits_{i=1}^n \biggl( \sum\nolimits_{j=1}^n \log \hat{\pi}_{ij}^{(4)}(\beta^{\rm sim}) - \log \hat{\pi}_i^{(1)}(\beta^{\rm obs})\biggr) \notag\\
& \quad\quad\quad - \frac{1}{2} \bigl\Vert \beta^{\rm sim} - \beta^{\rm obs}\bigr\Vert^2, \label{Eq2.20}
\end{align}
where such a learned summary statistic will serve as an approximation to the exact log-likelihood function for the Bayesian inference.

Here, we remark that the transformation of data-to-learned summary statistics under conditional moment restrictions discussed above can be extended for an inference task over certain sample sets of the observation data that are considered as an empirical relevance or believed to be particular importance. In what follows, suppose $\mathcal{S}$ is a sample set from the observation data $y^{\rm obs} = \bigl\{y_i^{\rm obs}\bigr\}_{i=1}^n$, i.e., $\mathcal{S} \subseteq y^{\rm obs}$. Then, we consider the following optimization problem
\begin{align}
\min_{\pi_{ij}, 1\le i,j \le n; \, \beta^{\rm sim}} &\, \frac{1}{\gamma(\gamma +1)} \sum\nolimits_{i=1}^n \chi(y_i^{\rm obs} \in \mathcal{S}) \sum\nolimits_{j=1}^n \left[w_{ij}\biggl(\frac{\pi_{ij}} {w_{ij}} \biggr)^{\gamma +1} - 1 \right],  \label{Eq2.21}\\
&\text{s.t.} \notag\\
& \quad \sum\nolimits_{j=1}^n \pi_{ij} g\bigl(y_{\theta,j}^{\rm sim}, \beta^{\rm sim} \bigr) = 0, \notag\\
& \quad \sum\nolimits_{j=1}^n \pi_{ij} = 1, \quad i=1, 2, \ldots, n, \notag
\end{align}
with $\pi_{ij} \ge 1$ for $1 \le i,j \le n$, while the indicator function $\chi (y_i^{\rm obs} \in \mathcal{S})$, for $i=1$, $2$, \ldots, $n$, will be $1$, if $y_i^{\rm obs}$, belongs to $\mathcal{S}$, otherwise, it is zero (see also Equation~\eqref{Eq2.15}). Note that the solutions $\hat{\pi}_{ij}^{(5)}(\beta^{\rm sim})$, with $1 \le i,j \le n$, for the above optimization problem will lead to a data-to-learned summary statistic $\ell_{\rm learned}\bigl(\theta \vert y^{\rm obs}\bigr)$ (which is based on the empirical log-likelihood ratio between the local Cressie-Read discrepancy contrast probabilities $\hat{\pi}_{ij}^{(5)}(\beta^{\rm sim})$, for $1 \le i,j \le n$, and $\bigl\{\hat{\pi}_i^{(1)}\bigl(\beta^{\rm obs}\bigr)\bigr\}_{i=1}^n$ as well as the Euclidean norm distance between the estimated parameters $\beta^{\rm sim}$ and $\beta^{\rm obs}$ associated with the conditional moment restrictions), that is
\begin{align}
\ell_{\rm learned}\bigl(\theta \vert y^{\rm obs}\bigr) =& \sum\nolimits_{i=1}^n \biggl( \sum\nolimits_{j=1}^n \log \hat{\pi}_{ij}^{(5)}(\beta^{\rm sim}) - \log \hat{\pi}_i^{(1)}(\beta^{\rm obs})\biggr) \notag\\
& \quad\quad\quad - \frac{1}{2} \bigl\Vert \beta^{\rm sim} - \beta^{\rm obs}\bigr\Vert^2, \label{Eq2.22}
\end{align}
where such a data-to-learned summary statistic is also expected to serve as a proxy to the exact log-likelihood function for the Bayesian inference.

\subsection{Weakly dependent observation data with block-wise implementations} 
Suppose that the observation data $y^{\rm obs} = \bigl\{y_i^{\rm obs}\bigr\}_{i=1}^n$ are weakly dependent time series observations. In what follows, we form dataset blocks as follows: 
\begin{enumerate}[$\quad\quad$]
\item  For an integer $m$ (with $m < n$), the $k^{\rm th}$-block of observation data $B_k^{\rm obs}$ is given by
\begin{align}
 B_k^{\rm obs} = \bigl( y_k^{\rm obs}, \, y_{k+1}^{\rm obs}, \ldots, y_{k+m-1}^{\rm obs} \bigr), \quad k = 1, \ldots, (n-m+1), \label{Eq2.23}
\end{align}
where we assume as $n \to \infty $, then $m \to \infty $ and $m = {\rm o}(\sqrt{n})$.
\end{enumerate}
Note that the purpose of blocking is to retain the dependence patterns of the $y_k^{\rm obs}$'s in each of the blocks $B_k^{\rm obs}$ of length $m$, with $k = 1$, \ldots, $(n-m+1)$. That is, as the sample size $n$ grows, if $m$ grows slowly, then it will capture information about the weak dependence in the observation data asymptotically in a fully nonparametric ways. Moreover, we can compute the $k^{\rm th}$ smoothed aggregated (moment) function as follows
\begin{align}
 \psi \bigl(B_k^{\rm obs}, \beta^{\rm obs} \bigr) = \frac{1}{m} \sum\nolimits_{s=0}^{m-1} g\bigl(y_{k+s}^{\rm obs}, \beta^{\rm obs} \bigr), \quad k = 1, \ldots, (n-m+1). \label{Eq2.24}
\end{align}
Note that if we proceed by considering the following optimization problem - based on the Cressie-Read discrepancy criterion - of the form
\begin{align}
\min_{\pi_k, 1\le k \le n-m+1; \, \beta^{\rm obs}} &\, \frac{1}{\gamma(\gamma +1)} \sum\nolimits_{k=1}^{n-m+1} \left[\bigl((n-m+1) \pi_k \bigr)^{\gamma +1} - 1 \right], \label{Eq2.25}\\
&\text{s.t.} \notag\\
& \quad \sum\nolimits_{k=1}^{n-m+1} \pi_k \psi\bigl(B_k^{\rm obs}, \beta^{\rm obs} \bigr) = 0, \notag\\
& \quad \sum\nolimits_{k=1}^{n-m+1} \pi_k = 1, \notag
\end{align}
with $\pi_k \ge 1$ for $k=1$, \ldots, $(n-m+1)$ (see Equation~\eqref{Eq2.1}). Then, by following the same arguments as in Subsection~\ref{S2.1}, for fixed $\beta^{\rm obs}$, the solutions for $\hat{\pi}_k^{(6)}(\beta^{\rm obs})$, for $k=1$, \ldots, $(n-m+1)$, satisfy the following system of equations
\begin{align}
\frac{(n-m+1)}{\gamma} \bigl((n-m+1) \hat{\pi}_k^{(6)}(\beta^{\rm obs}) \bigr)^{\gamma} - \hat{\lambda}(\beta^{\rm obs})^T g\bigl(y_i^{\rm obs}, \beta^{\rm obs} \bigr) - \hat{\eta}(\beta^{\rm obs}) = 0, \label{Eq2.26}
\end{align}
where $\hat{\lambda}(\beta^{\rm obs})$ and $\hat{\eta}(\beta^{\rm obs})$ are, respectively, the Lagrange multipliers associated with the constraints 
\begin{align*}
\sum\nolimits_{k=1}^{n-m+1} \hat{\pi}_k^{(6)}(\beta^{\rm obs}) \psi\bigl(B_k^{\rm obs}, \beta^{\rm obs} \bigr) = 0 \quad \text{and} \quad \sum\nolimits_{k=1}^{n-m+1} \hat{\pi}_k^{(6)}(\beta^{\rm obs}) = 1.
\end{align*}
Similarly, for the simulation outputs $y_{\theta}^{\rm sim} =\bigl\{y_{\theta,i}^{\rm sim} \bigr\}_{i=1}^n$ with similar dataset blocks, i.e.,
\begin{align}
 B_k^{\rm sim} = \bigl( y_{k,\theta}^{\rm sim}, \, y_{k+1,\theta}^{\rm sim}, \ldots, y_{k+m-1, \theta}^{\rm sim} \bigr), \quad k = 1, \ldots, (n-m+1), \label{Eq2.27}
\end{align}
the corresponding Cressie-Read discrepancy contrast probabilities $\hat{\pi}_k^{(7)}(\beta^{\rm sim})$, for $k=1$, \ldots, $(n-m+1)$. Moreover, we can introduce the data-to-learned summary statistics $\ell_{\rm learned}\bigl(\theta \vert y^{\rm obs}\bigr)$, that serves as an approximation to the exact log-likelihood function in the same way as in Subsection~\ref{S2.1} as follows
\begin{align}
\ell_{\rm learned}\bigl(\theta \vert y^{\rm obs}\bigr) = \sum\nolimits_{k=1}^{n-m+1} \biggl( \log \hat{\pi}_k^{(7)}(\beta^{\rm sim})  - \log \hat{\pi}_k^{(6)}(\beta^{\rm obs})\biggr) - \frac{1}{2} \bigl\Vert \beta^{\rm sim} - \beta^{\rm obs}\bigr\Vert^2. \label{Eq2.28}
\end{align}

\section{Concluding remarks} \label{S3}
 In this paper, which is Part~1 of a two-part paper series, we considered a simulation-based inference with learned summary statistics, where such a learned summary statistics serves as an empirical-likelihood in the Bayesian setting, when the exact likelihood function associated with the observation data and the simulation model is difficult to obtain in a closed form or computationally intractable. In particular, we used a transformation technique that leverages the Cressie-Read discrepancy measures under moment (and also conditional) restrictions models for summarizing the learned statistics between the observation data and the simulation outputs. Here, we stressed that such a data-to-learned summary statistic, which also preserve the statistical power of the inference, serves as a proxy to the exact log-likelihood function (i.e., an approximation to the exact intractable log-likelihood). Moreover, such a transformation of data-to-learned summary statistics allows the simulation outputs to be conditioned on the observation data, so that the inference task can be performed over certain sample sets of the observation data that are considered as an empirical relevance or believed to be particular importance. Here, our intent is to provide step-by-step coherent mathematical arguments that are necessary for the exposition of simulation-based inference with learned summary statistics, rather than considering any specific numerical problem. Although, some numerical works using the proposed computational frameworks have been done and detailed results will be presented elsewhere (i.e., in Part~2 of this two-part series paper). Finally, it is worth remarking that such an inference framework is suitable for implementation in distributed computing, i.e., computational tasks involving both the data-to-learned summary statistics and the Bayesian inferencing problem can be posed as a unified distributed inference problem that will exploit distributed optimization and MCMC algorithms for supporting large datasets associated with complex simulation models.

\end{document}